\documentclass[letterpaper,10pt,journal,twoside]{IEEEtran}  
\IEEEoverridecommandlockouts                 
\usepackage[font=footnotesize]{caption}  
\def\name{TeachingBot}   
\usepackage[colorlinks,linkcolor=blue,anchorcolor=blue,citecolor=blue]{hyperref} 

\usepackage{array} 
\usepackage{cancel} 
\usepackage{multicol}
\usepackage[numbers]{natbib} 

\usepackage{algorithmic}    
\usepackage{algorithm}  
\usepackage{epsf} 

\usepackage{graphicx} 
\usepackage{color}
\usepackage{amssymb} 
\usepackage{wrapfig} 
\usepackage{mathtools} 
\usepackage{bbm}  
\usepackage{amsmath}   
\usepackage{color}  
\usepackage{subfigure,bbm} 

\usepackage{xspace}  

\usepackage{multirow}  
\usepackage{booktabs}  

\usepackage{CJKutf8}

\newcommand{\figref}[1]{Fig.~\ref{#1}}

\newcommand{\secref}[1]{Section~\ref{#1}}
\newcommand{\algref}[1]{Algorithm~\ref{#1}}

\def\mbB{\mathbf{B}}
\def\mbC{\mathbf{C}}
\def\mbD{\mathbf{D}}

\def\mbS{\mathbf{S}}
 
\def\mbK{\mathbf{K}}

\def\md{\mathcal{D}}

\def\mn{\mathcal{N}}

\def\mp{\mathcal{P}}

\def\bx{\boldsymbol{x}}  
  
\def\bs{\boldsymbol{s}}  
  
\def\bk{\boldsymbol{k}} 
\def\bt{\boldsymbol{t}} 
\def\bu{\boldsymbol{u}}

\def\bc{\boldsymbol{c}}

\def\bK{\boldsymbol{K}}

\def\bF{\boldsymbol{F}}

\def\bK{\boldsymbol{K}}

\def\bJ{\boldsymbol{J}}

\def\bxi{\boldsymbol{\xi}}

\def\bmu{\boldsymbol{\mu}} 
\def\bSigma{\boldsymbol{\Sigma}} 
\def\bOmega{\boldsymbol{\Omega}}

\def\bchi{\boldsymbol{\chi}}

\def\bGamma{\boldsymbol{\Gamma}}

\def\mbK{\boldsymbol{\mathcal{K}}}

\newcommand{\ea}[1]{\begin{equation}\begin{aligned}\centering \small #1 \end{aligned}\end{equation}}

\def\md{\mathcal{D}}

\def\mn{\mathcal{N}}

\def\mp{\mathcal{P}}

\def\bx{\boldsymbol{x}}  
  
\def\bs{\boldsymbol{s}}  
  
\def\bk{\boldsymbol{k}} 
\def\bt{\boldsymbol{t}} 
\def\bu{\boldsymbol{u}}

\def\bc{\boldsymbol{c}}

\def\bK{\boldsymbol{K}}

\def\bF{\boldsymbol{F}}

\def\bK{\boldsymbol{K}}

\def\bJ{\boldsymbol{J}}

\def\mbB{\boldsymbol{\mathcal{B}}}  
\def\mbC{\boldsymbol{\mathcal{C}}}  
\def\mbD{\boldsymbol{\mathcal{D}}}

\def\mbS{\boldsymbol{\mathcal{S}}}  
   
\def\mbK{\boldsymbol{\mathcal{K}}}

\def\bxi{\boldsymbol{\xi}}

\def\bmu{\boldsymbol{\mu}} 
\def\bSigma{\boldsymbol{\Sigma}} 
\def\bOmega{\boldsymbol{\Omega}}

\def\bchi{\boldsymbol{\chi}}

\def\bGamma{\boldsymbol{\Gamma}} 
   
\def\btraj{\boldsymbol{\mathcal{T}}}

\def\bGP{\boldsymbol{\mathcal{GP}}}

\definecolor{darkgreen}{rgb}{0.0, 0.5, 0.0}

\title{TeachingBot: Robot Teacher for Human Handwriting}      
\author{
Zhimin Hou\textsuperscript{\textnormal{1}}$^{*}$, 
Cunjun Yu\textsuperscript{\textnormal{2}}$^{*}$, 
David Hsu$^{2, 3}$, 
Haoyong Yu$^{1}$
}

\begin{document}  
\begin{CJK*}{UTF8}{gbsn}  
\maketitle

\let\thefootnote\relax\footnotetext{
* Equal contribution. \\
\indent The authors are associated with the $^{1}$Department of Biomedical Engineering, $^{2}$School of Computing, $^{3}$Smart System Institute, National University of Singapore. 
\\ 
\indent In Section~\ref{exp}, all subjects gave informed consent and testing was approved by National University of Singapore Institutional Review Board (NUS-IRB-2023-875). 
} 

\begin{abstract}  
Teaching and learning physical skills often require one-on-one interaction, making it difficult to scale up, as there are not enough human teachers. Robots offer an attractive alternative. This paper presents \textit{\name{}}, an adaptive robotic system that teaches handwriting to human learners through physical interaction. Robot teaching poses two major challenges: (i) adapting to the individual handwriting style of the learner and (ii) maintaining an engaging learning experience. For the first challenge, \name{} uses a probabilistic model to capture the learner’s writing style from their writing samples. Drawing on the insight that effective teaching  balances standardization with individuality, the system generates a personalized teaching trajectory that aligns with the learner’s natural writing. For the second challenge, \name{} employs variable impedance control to guide the learner, dynamically adjusting the strength of physical guidance based on the learner’s performance. Human-subject experiments with 15 participants demonstrate the effectiveness of \name{}, showing clear improvement in learners' handwriting and engagement over baseline methods. 
\end{abstract}

\IEEEpeerreviewmaketitle    

\vspace{-0.4cm}  
\section{Introduction}\label{sec-introduction} 
Robots play a crucial role in various physical interaction tasks with humans~\cite{losey2018review}. While robots have traditionally served as learners~\cite{arduengo2021task,zhou2019learning}, collaborators~\cite{losey2022learning}, or assistants~\cite{pehlivan2015minimal}, they hold great potential as teachers for humans~\cite{yu2022coach}. \emph{How can robots provide effective, personalized physical guidance for teaching physical skills when human teachers are scarce?}
Specifically, this work focuses on handwriting, an essential physical skill in daily life. The robot serves as the \textit{teacher}, providing guidance through physical interaction with humans to teach handwriting. 

Teaching humans to write effectively presents two key challenges. First, human learners exhibit diverse writing styles for the same character~\cite{lemaignan2016learning,luo2024callirewrite,wang2024rodal}, with variations in stroke length, position, and direction—especially evident in Chinese characters~\cite{luo2024callirewrite}. Learners naturally prefer guidance that accommodates their individual writing styles. Second, determining the appropriate level of guidance is difficult: insufficient support fails to facilitate learning, while excessive assistance can create dependency~\cite{pehlivan2015minimal} and hinder skill acquisition~\cite{yu2022coach}. 
A successful robot teaching system must therefore (1) adapt to individual writing styles and (2) provide an appropriate level of physical guidance that promotes active engagement. 

\begin{figure}[!t]  
\setlength{\abovecaptionskip}{0.1cm}  
\centering
\includegraphics[width=1.0\columnwidth]{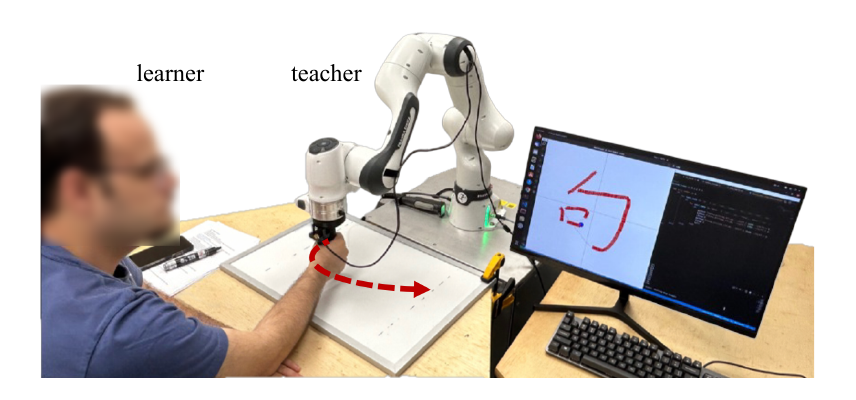}
\caption{\textit{\name{}}: the robot teacher guides the learner in writing Chinese characters, adapting to the learner’s preferences and proficiency.}    
\label{fig:introduction}     
\vspace{-15pt}  
\end{figure}

We present \textbf{\textit{\name{}}}, an adaptive robot teaching system designed to teach Chinese character writing through physical guidance, as illustrated in Fig.~\ref{fig:introduction}. Our key insight is that effective teaching should balance standardization with individuality since guidance is more effective when grounded in the human learner’s own writing style. \name{} achieves this through a two-step adaptive approach.

First, the system captures the human learner’s writing style by collecting their writing trajectories and modeling them probabilistically. Drawing inspiration from learning-from-demonstration methods~\cite{huang2018generalized,huang2019kernelized,jaquier2020learning}, we use a mixture of Gaussian distributions to encode the human learner’s writing style, capturing both the average writing trajectory and its variations. By combining this model with the reference character the robot aim to teach, we generate a teaching trajectory that respects the human learner’s writing preferences. 

Second, the robot executes the teaching trajectory through compliant physical interaction using variable impedance control~\cite{proietti2016upper,yang2022task,hou2025contextual}. The human learner holds the robotic arm’s end effector, which guides them through the generated teaching trajectory~(see Fig.~\ref{fig:introduction}). The system dynamically adjusts the guidance force: applying stronger correction when the human learner strays far from the reference, and reducing assistance when deviations are minor to encourage exploration and maintain engagement~\cite{pehlivan2015minimal,pareek2023ar3n,li2018iterative}. 
Through this adaptive strategy, \name{} creates a personalized, engaging learning experience that respects individual writing styles while guiding learners toward reference character style. 
This approach is especially suitable in scenarios where detailed models of human learning dynamics are unavailable~\cite{srivastava2022assistive,yu2022coach}.

It is crucial to distinguish between \textit{robot learning} and \textit{robot teaching}. While robot learning focuses on robots acquiring new capabilities, robot teaching, as exemplified by \name{}, positions robots as instructors for human learners. Through human-subject experiments\footnote{IRB has been acquired} with 15 participants with varying levels of Chinese writing proficiency, we demonstrate that \name{} significantly improves handwriting quality compared to two baseline methods. Participants showed greater accuracy in replicating both the overall structure and fine-grained stroke details of characters. Additionally, increased interaction forces during training indicate enhanced learner engagement, highlighting the effectiveness of adaptive physical guidance in promoting active skill acquisition. 

One limitation of this work is that it does not explicitly address long-term retention or how well the acquired skills transfer beyond the immediate training sessions.

\vspace{-0.1cm} 
\section{Related Work}
\subsection{Teaching Algorithms}  
Various algorithms have been used for human learning, with successes in areas such as crowd classification and concept acquisition~\cite{Singla2014,Zilles2011,Doliwa2010,Aodha_2018_CVPR}. However, mastering intricate motor control skills through mere visual or verbal cues remains a challenge. 
Recent advances integrate skill discovery techniques from reinforcement learning to establish curricula based on skill decomposition, enhancing human motor skill acquisition~\cite{srivastava2022assistive}. The emergence of advanced language models has introduced language correction as a tool to facilitate human learning~\cite{srivastava2023generating}. 
The concept of robot teaching has emerged in the realm of robot-assisted learning~\cite{hood2015children,lemaignan2016learning}. Robots can physically facilitate human learning by using specified dynamics~\cite{yu2022coach,Tian2023TowardsMA,pomdpteaching}. 
In this work, instead of only providing physical guidance, we aim to achieve adaptive teaching by learning and incorporating the style of the learner in the physical guide. 
\vspace{-8pt} 
\subsection{Robot-assisted Training} 
While social robots have successfully facilitated language learning and writing skills~\cite{cowriter}, physical human-robot interaction~(pHRI) proves more effective for tasks requiring subtle motor control by enabling direct physical guidance~\cite{hou2025contextual}. 
Robot-assisted training has been extensively studied in various rehabilitation applications~\cite{proietti2016upper}. 
The key challenges in pHRI-based learning systems lie in trajectory generation and interaction control. For trajectory generation, probabilistic methods like Gaussian Mixture Models (GMM)/Regression (GMR) and Dynamic Movement Primitives (DMP)~\cite{huang2018generalized,huang2019kernelized,pastor2009learning} enable robots to generate smooth, human-like reference trajectories. 
Advanced approaches such as Probabilistic DMP~(ProMP) and Gaussian Process~(GP) models~\cite{paraschos2018using,jaquier2020learning,zhou2019learning} further improve adaptability through basis function representations. 
While these methods have proven effective in robot-assisted rehabilitation~\cite{proietti2016upper,maaref2016bicycle}, they primarily focus on robots learning from experts/therapists rather than actively optimizing trajectory generation for human skill acquisition. 

The effectiveness of physical guidance also depends critically on the adjustment of the interaction intensity~\cite{proietti2016upper}. 
Research shows that active user engagement enhances neural plasticity and learning efficiency~\cite{warraich2010neural,pehlivan2015minimal}. 
Variable impedance controllers have emerged as key tools for modulating this interaction~\cite{caldarelli2022perturbation,proietti2016upper}, with approaches ranging from error-based parameter adjustment~\cite{pehlivan2015minimal,yang2022task} to model-free learning-based methods~\cite{pareek2023ar3n,li2018iterative,hou2022hierarchical}. 
However, these approaches have only shown effectiveness in promoting active participation, while failing to simultaneously accommodate individual learning preferences and encourage user participation during interaction control. 
\name{} addresses these challenges by dynamically adjusting the teaching intensity based on both the learner's style and the real-time writing performance. 
\begin{figure*}[!t]
\setlength{\abovecaptionskip}{-0.00cm}    
\centering
\includegraphics[width=0.95\textwidth]{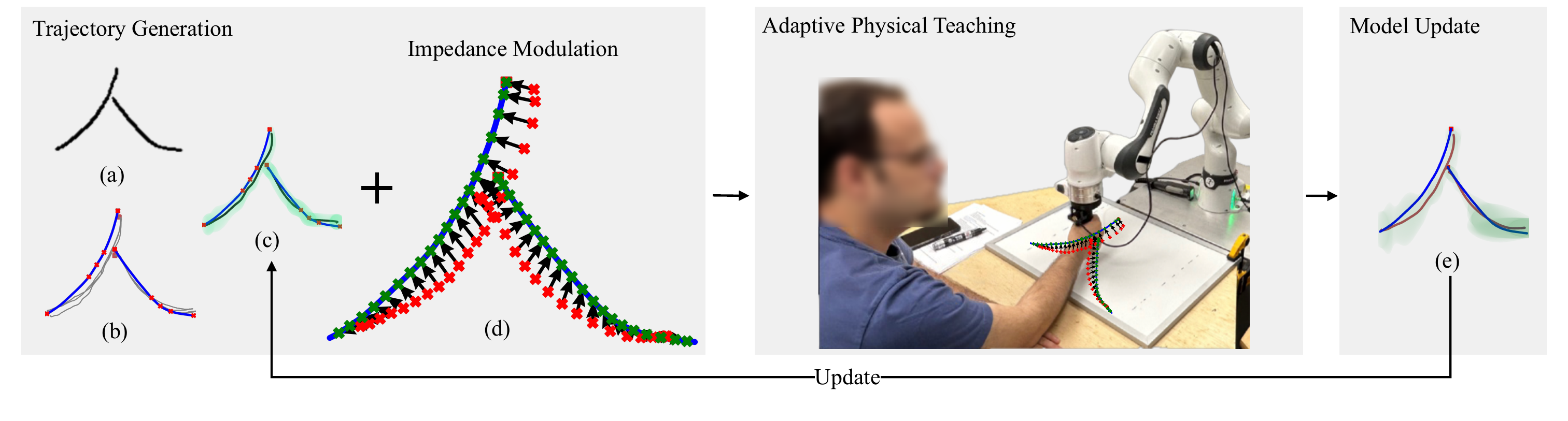} 
\caption{
{Overview of \name{}. } First, a reference character  (a) is selected, and waypoints (b) are extracted from both the reference and the learner’s handwriting. Their differences guide initial impedance modulation (d), where the \textcolor{blue}{blue}/\textcolor{darkgreen}{green} path shows the reference trajectory/waypoints, the \textcolor{red}{red} cross indicates the learner’s mean writing waypoints, and black arrows represent the potential correction forces used to modulate the impedance parameters for robot control. The robot then physically guides and teaches the learner. After teaching, a GMM-GMR (e) updates the learner’s writing style, and GMR-GP generates a new training trajectory for the next teaching iteration (black in c). 
}    
\label{fig:raw_overview}  
\vspace{-12pt}    
\end{figure*}

\vspace{-0.1cm} 
\section{Methodology}\label{sec-methods}    
\subsection{Problem Formulation}
We build on prior work~\cite{yu2022coach,srivastava2022assistive} that models the \textit{target task} (the skill to be taught) as a Markov Decision Process~(MDP) and the \textit{teaching task} (the teaching process) as a Partially Observable MDP~(POMDP). Here, the target task is writing Chinese characters, and the teaching task is to guide humans in learning this skill. The \textit{teaching policy} influences learner performance through interactive actions, such as generating reference trajectories and adjusting physical interaction impedance.  The objective of \name{} is to adaptively teach learners to write the reference character.  Defining the POMDP elements is challenging due to the lack of an accurate human learning model, so we adopt a simplified solution rather than solving the full POMDP.
\name{} adjusts its support to the human learner's needs purely based on the current observation of the human learner's writing style encoded by the probabilistic learning method GMM/GMR and writing performance. 

Building upon the formulation for pHRI~\cite{pehlivan2015minimal}, the dynamics of the \textit{robot teacher} are expressed in the end-effector space as 
\ea{ 
\label{equ_dynamics} 
\mathcal{\boldsymbol{M}}(\bx)\ddot{\bx} + \mathcal{\boldsymbol{C}}(\bx, \dot{\bx})\dot{\bx} + \mathcal{\boldsymbol{G}}(\bx) = \bF_h + \bF_r 
}  
where $\bx$ and $\dot{\bx}$ are the actual position and velocity, respectively. $\bF_r$ is the assistive force provided by \textit{the robot teacher}. $\bF_h$ is the \textit{human learner} actively applied force, typically measured to evaluate the engagement in pHRI tasks~\cite{pehlivan2015minimal,hou2025contextual,li2018iterative}. 
A variable impedance controller $\pi(\boldsymbol{u} |\bx, \dot{\bx}; \bx_d, \dot{\bx}_d, \boldsymbol{\mathcal{K}}_d,\boldsymbol{\mathcal{B}}_d)$ is implemented to provide corrective guidance given the reference trajectory and impedance parameters~\cite{hou2025contextual}. 
$\bx_d \in \mathbb{R}^3$ and $\dot{\bx}_d \in \mathbb{R}^3$ are the reference position and velocity, respectively. 
At each actuation step with a sample interval $T_s$, the control input $\bu$ is designed as 
\ea{
\label{equ_impedance_command} 
\bu =& \bJ^{T}\big[- \boldsymbol{\mathcal{K}}_d(\bx - \bx_d) -\left.\boldsymbol{\mathcal{B}}_d(\dot{\bx} - \dot{\bx}_d)\right. + \bF_{fd} \big], 
} 
where $\bJ$ is the Jacobian matrix. 
$\bF_{fd}$ is introduced to compensate for the robot dynamics as formulated in (\ref{equ_dynamics}). 
$\boldsymbol{\mathcal{K}}_d \in \mathbb{R}^{3 \times 3}$ and $\boldsymbol{\mathcal{B}}_d \in \mathbb{R}^{3 \times 3}$ are the  reference stiffness and damping. They are regulated to adjust the assistance level provided by the \textit{robot teacher} and the engagement level of the \textit{human learner}~\cite{proietti2016upper,yang2022task,li2018iterative}. 
\vspace{-0.2cm}   
\subsection{Overview of TeachingBot}  
The overview of \name{} is depicted in~\figref{fig:raw_overview}. 
The reference character is denoted by the variable $\bc \in \mbC$. $\bs_i \in \mbS_{\bc}$ represents $i$-th stroke of character $\bc$. 
The dataset of reference Chinese characters, $\mbC$, includes images of each character $\{\boldsymbol{I}_c\}$ for the human users to learn~(see \figref{fig:raw_overview}(a)). $\mbS_{\bc}$ includes all stroke images of the character $\bc$. 
To facilitate the generation of robot teaching trajectories, reference waypoints $\{\bchi_{\bc}^n\}_{n=1}^{N}$ of all strokes are extracted from the image of the reference character based on image processing method, $N$ is the number of waypoints and $\bchi^n_c$ denotes the $n$-th reference waypoint. 
As illustrated in~\figref{fig:raw_overview}(b), the reference waypoints of two strokes are extracted and plotted as the blue lines. 
Given a reference character $\bc$, our TeachingBot is implemented following three phases, as presented in Algorithm~\ref{alg-training-procedure}. 
During the pre-test and evaluation phases, the human learner is instructed to write the reference character for $L$ iterations without the physical guidance of the robot. 
\begin{algorithm}[t]  
\footnotesize
\caption{\footnotesize\name{}}          
\label{alg-training-procedure}   
\begin{algorithmic}[1]
\STATE \textbf{Input:} Character set $\mbC$, number of teaching iterations $M$, number of evaluation iterations $L$  
\STATE \textbf{Initialize:} $\mbD_V^0 \leftarrow \emptyset$, $\mbD_L^0 \leftarrow \emptyset $, $\mbK_s^0 \leftarrow \mathbf{0}$, $\mbB_s^0 \leftarrow \mathbf{0}$, GMR-GP
\FOR{each character $\bc \in \mbC$}  
    \STATE Extract reference waypoints $\bchi_c$ from image $\boldsymbol{I}_c$
    \STATE \textbf{/// Pre-test Phase ///}
    \FOR{$l = 1$ to $L$}
        \STATE Collect learner's trial image $\boldsymbol{I}_a^l$ and extract waypoints $\bchi_a^l$
    \ENDFOR
    \STATE Encode learner’s initial writing style $\overline{\bchi}_a^I$ from $\{\bchi_a^l\}_{l=1}^L$
    \STATE Initialize robot impedance parameters $\mbK_r$, $\mbB_r$
    \STATE \textbf{/// Teaching Phase ///}
    \FOR{$m = 1$ to $M$}
        \FOR{each stroke $\bs \in \mbS_{\bc}$}
            \STATE Update learner’s writing style from $\mbD_L^{m-1}$
            \STATE Retrieve $H$ training via-points from $\mbD_V^m$
            \STATE Generate teaching trajectory $\bchi_d^m$ via GMR-GP model
            \STATE Compute impedance: stiffness $\mbK_d^m$, damping $\mbB_d^m$
            \STATE Apply physical guidance using VIC (Eq.~\ref{equ_impedance_command})
            \STATE Record learner’s actual writing trajectory $\bchi_a^m$ 
            \STATE Update dataset: $\mbD_L^m \leftarrow \mbD_L^{m-1} \cup \{\bt, \bchi_a^m\}$, $\mbD_V^m \leftarrow \emptyset$
        \ENDFOR
    \ENDFOR
    \STATE \textbf{/// Evaluation Phase ///}
    \FOR{$l = 1$ to $L$}
        \STATE Collect evaluation image $\boldsymbol{I}_a^l$ and extract waypoints $\bchi_a^l$ 
    \ENDFOR
    \STATE Encode learner’s writing style $\overline{\bchi}_a^E$ from $\{\bchi_a^l\}_{l=1}^L$ 
\ENDFOR
\end{algorithmic}  
\end{algorithm}  
 
During the teaching phase, a complete round of writing the given character is defined as one \textit{teaching iteration}. Robot teaching is repeated for $M$ iterations, at $m$-th teaching iteration~($m \in M$), the previous $L$ actual writing trajectories are collected as $\{\btraj^l_a\}_{l=1}^L$, which are downsampled into actual writing waypoints $\{\{\bchi^{l,n}_a\}_{n=1}^{N}\}_{l=1}^L$. 
A dataset comprising all time-driven writing waypoints, $\md_L^m=\{\{\bt_n^l,~\bchi^{l,n}_a\}_{n=1}^{N}\}_{l=1}^L$, is then constructed to encode the human learner's writing style. 
Afterwards, the key via-points are extracted from the reference waypoints and stored in $\md_V^m$, defined as training via-points. 
A GMR-GP model, $f(\bchi_r|\bt,\md_L^m,\md_V^m)$, is trained to generate training waypoints $\{\bchi_d^{m,n}\}_{n=1}^N$ based on the encoded human learner's writing style and the training via-points. 
Finally, the training waypoints are interpolated to create the reference trajectory integrating with the adapted impedance parameters for physical teaching interactive control~(see Section~\ref{subsec-physical-interactive-control}). 
As illustrated in~\figref{fig:raw_overview}, the human learner would grasp the robot's handle, and the robot teacher guides the human learner to write the reference character by following a generated reference trajectory. 
Afterwards, the reference trajectory and impedance parameters will be updated for the robot teacher to personalize the teaching task. 
\vspace{-0.2cm}   
\subsection{Human Learner Writing Style Learning}\label{writing-style-evaluation} 
For the $m$-th teaching iteration, the human learner's writing style is encoded using GMM/GMR with $Z$ components. 
The joint distribution of the input time $\bxi_i = \bt \in \mathbb{R}^{d_i}$, and the output writing waypoint $\bxi_o = \bchi \in \mathbb{R}^{d_o}$ is modelled as 
\ea{ 
\label{equ-gmr-model-prior}  
\mp(\boldsymbol{\xi}) = \sum\nolimits_{z=1}^Z h_z\mn(\bxi; \bmu_z, \bSigma_z), \bxi =[\bxi_i,\bxi_o]^T,
} 
where $h_z$, $\bmu_z$, and $\bSigma_z$ are the prior probability, mean, and covariance of the $z$-th Gaussian component, respectively. 
These parameters are optimized using the Expectation-Maximization algorithm given the dataset $\md_L^{m-1}$\cite{huang2019kernelized,jaquier2020learning}. 
Therefore, the writing style of human learner is represented by a probabilistic writing waypoints $\{\widehat{\bchi}_a^{n}\}_{n=1}^N$. 
Each waypoint $\widehat{\bchi}_a^n$ is retrieved from a conditional Gaussian distribution $\mp(\widehat{\bchi}_a^n |\bt_n) = \mn(\widehat{\bmu}({\bt_n}), \widehat{\bSigma}({\bt_n}))$. 
$\widehat{\bmu}({\bt_n})$ and $\widehat{\bSigma}({\bt_n})$ are the conditional mean and covariance, respectively. 
The covariance $\widehat{\bSigma}({\bt_n})$ encapsulates the variability of the human learner's potential writing trajectories, providing a measure of uncertainty and flexibility in their writing style. 

As depicted in \figref{fig:raw_overview}(a) and \figref{fig:raw_overview}(b), $L$=3 writing waypoints of the reference character $\bc$ are represented by gray lines. 
The GMMs, with $Z$=8 components, are optimized and visualized as the green ellipses in \figref{fig:raw_overview}(e). 
The mean writing waypoints and variance, derived from GMMs, are depicted as the green line and green shaded region in \figref{fig:raw_overview}(f). 
\subsection{Training Via-points Extraction}     
The curvature-based trajectory compression is employed to extract the training via-points from the reference waypoints $\bchi_c$. 
Particularly, we retain waypoints where the curvature (change in direction) is highest, as these waypoints are likely to represent features of the reference character. 
Given the reference waypoints $\{\bchi_{\bc}^n\}_{n=1}^{N}$, calculate curvature \(\kappa_n\) for each interior waypoint \(\bchi_{\bc}^n\) using the first order derivative $\dot{\bchi}^n_{\bc}$ and second order derivative $\ddot{\bchi}^n_{\bc}$. 
The curvature can be computed as \(\kappa_n = \ddot{\bchi}^n_{\bc} / (1+(\dot{\bchi}^n_{\bc})^2)^{1.5}\). 
Then, $H$ waypoints with the highest curvature values are selected as training via-points, indicating significant changes in trajectory direction. As introduced in~\algref{alg-training-procedure}, during the $m$-th teaching iteration, the training via-points are obtained and stored in $\md_V^m = \{\bt_h^m, \bchi_c^h\}_{h=1}^H$~($H<<N$). As depicted in Fig.~\ref{fig:raw_overview}(b), the extracted training via-points~($H=5$) are visualized as red scatter points. 

\subsection{Training Waypoints Generation}\label{subsec-gmr-gp} 
For the $m$-th teaching iteration, given the previous writing waypoints in $\md_L^{m-1}$, the training waypoints need to be generated to maintain the current writing style and follow the instruction of training waypoints in $\md_V^m$. 
Multi-output GP~(MOGP) is employed to fit the deterministic relationship $\bxi_o = f(\bxi_i) + \boldsymbol{\epsilon}_t$ from the input $\bxi_i = \bt$ to vector-valued output $\bxi_o = \bchi$, $\boldsymbol{\epsilon}_t = [\boldsymbol{\epsilon}_t^1, \boldsymbol{\epsilon}_t^p, \cdots, \boldsymbol{\epsilon}_t^{d_o}]$,  $\boldsymbol{\epsilon}_t^p \sim \mn(0, \sigma_p^2)$. 
The distribution of $\bchi$ input $\bt$ is given by
$\bchi(\bt) \sim \bGP(\bmu(\bt),\bk(\bt, \bt^{\prime}))$, 
where $\bmu(\cdot) : \mathbb{R}^{d_i} \rightarrow \mathbb{R}^{d_o}$ and $\bk(\cdot, \cdot) : \mathbb{R}^{d_i} \times \mathbb{R}^{d_i} \rightarrow \mathbb{R}^{d_o} \times \mathbb{R}^{d_o}$ are the mean and kernel function. 
The joint distribution of observed samples and the predicted output $\bchi^{*}$ of the input $\bt_{*}$ is given as follows  
\ea{ 
\begin{bmatrix} 
\bchi^{1:N}\\
\bchi^{*}    
\end{bmatrix}  
\sim
\mn
\left(
\begin{array}{cc}   
\begin{bmatrix}   
\bmu(\bt_{1:N}) \\  
\bmu(\bt_{*})   
\end{bmatrix} , &  
\begin{bmatrix}   
\bK(\bt,\bt) + \boldsymbol{\Sigma}_{\epsilon} & \bK(\bt, \bt_*) \\
\bK(\bt_*, \bt) &  \bK(\bt_*, \bt_*)    
\end{bmatrix}   
\end{array}     
\right),    
} 
where $\bchi^{1:N}$ are the observed values correspond to the input $\bt_{1:N}$. 
$\bK(\bt, \bt) \in \mathbb{R}^{Nd_o \times Nd_o}$, $\bK(\bt, \bt_*) \in \mathbb{R}^{Nd_o \times d_o}$, $\bK(\bt_*, \bt) \in \mathbb{R}^{d_o \times Nd_o}$, and $\bK(\bt_*, \bt_*) \in \mathbb{R}^{d_o \times d_o}$ are the Gram matrices that all elements are calculated using the kernel function over all input pairs $(\bt, \bt)$. 
Similarly to~\cite{jaquier2020learning,arduengo2021task}, the kernel function 
$\bk(\bt, \bt^{\prime}) = \sum\nolimits_{q=1}^Q \boldsymbol{\Xi}_q \bk_q(\bt, \bt^{\prime})$ is designed based on the Linear Model of Coregionalization~(LMC) assumption. 
$\boldsymbol{\Xi}_q \in \mathbb{R}^{d_o \times d_o}$ is a positive semi-definite coregionalization matrix. 
$\bk_q(\bt, \bt^{\prime})$ is the scalar kernel function. 

The posterior distribution of the human learner's writing waypoints is derived as a multivariate Gaussian distribution~(MGD), as 
$\mp(\bchi^{*} |\bt_{*}, \md_L^{m-1}) \sim \mathcal{N}(\bmu^{*}_L, \bSigma^{*}_L)$. 
The mean and covariance $\bmu^{*}_L$ and $\bSigma^{*}_L$ are computed as follows 
\ea{  
\label{equ-mean-variance-calculation}
\bmu^{*}_L &= \bmu(\bt_*) + \bK(\bt_*,\bt)(\bK(\bt,\bt) + \boldsymbol{\Sigma}_{\epsilon})^{-1}(\bchi^{1:N} - \bmu(\bt))\\  
\bSigma^{*}_L &= \bK(\bt_*,\bt_*) - \bK(\bt_*,\bt)(\bK(\bt,\bt) + \bSigma_{\epsilon})^{-1}\bK(\bt,\bt_*), 
}  
where the training via-points in $\md_V^m$ are treated as new observations to complement the previous observations in $\md_L^{m-1}$. 
Unlike standard MOGP, GMR-GP model replaces the prior mean value $\bmu(\bt_*)$ with the estimated GMR mean value. 
The kernel function $\bk(\bt, \bt^{\prime}) = \sum\nolimits_{z=1}^Z h_z(\bt)h_z(\bt^{\prime})\widehat{\bSigma}_z\bk_z(\bt,\bt^{\prime})$ is designed based on the learned variability. 
$h_z(\bt)$ and $\widehat{\bSigma}_z$ are the derived responsibilities and componentwise conditional covariance matrix calculated by (\ref{equ-gmr-model-prior}). 
The MGD for the training waypoints generation is rewritten as follows 
\ea{
\label{equ-final-distribution}
\mp(\bchi^{*}|\bt_{*}, \md_L^{m-1}, \md_V^m) &\propto \mp(\bchi^{*}|\bt_{*}, \md_L^{m-1}) \mp(\bchi^{*}|\bt_{*}, \md_V^m),\\
\mp(\bchi^{*}|\bt_{*}, \md_L^{m-1}, \md_V^m) &\sim \mn(\bmu_{L-V}^m, \bSigma_{L-V}^m), 
}  
where $\bmu_{L-V}^m$ and $\bSigma_{L-V}^m$ are derived based on $\{\bmu_L^*,\bSigma_L^*,\bmu_V^*,\bSigma_V^*\}$~\cite{arduengo2021task}. 
$\bmu_L^*$ and $\bSigma_L^*$ are the mean and covariance of $\mp(\bchi^{*}|\bt_{*}, \md_L^{m-1})$; 
$\bmu_V^*$ and $\bSigma_V^*$ are the mean and covariance of 
$\mp(\bchi^{*}|\bt_{*}, \md_V^m)$. 
Subsequently, the training waypoints $\{\bchi_d^{m,n}\}_{n=1}^N$ are sampled from the learned posterior distribution in (\ref{equ-final-distribution}) given $\{\bt_n\}_{n=1}^N$. 
Using the training via-points shown in \figref{fig:raw_overview}(b) and encoded writing styles in \figref{fig:raw_overview}(e), the sampled training waypoints are illustrated by the black line in \figref{fig:raw_overview}(c). 
\subsection{Physical Teaching Interactive Control}\label{subsec-physical-interactive-control} 
The actual writing trajectory and interaction force trajectory are respectively collected as $\btraj_a = [\bx(0), \bx(T_s), \dots, \bx(\Delta \bt_c)]$ and $\bGamma_a = [\bF_h(0), \bF_h(T_s), \dots, \bF_h(\Delta \bt_c)]$. $\Delta \bt_c$ is writing duration for the given reference character $\bc$. 
After obtaining the training waypoints $\{\bchi_d^{m,n}\}_{n=1}^N$ from the $m$-th teaching iteration, the reference trajectory $\btraj_d = [\bx_d(0), \bx_d(T_s), \cdots, \bx_d(\Delta \bt_c)]$ for the robot interactive control is generated by interpolation. 
The corresponding impedance $\boldsymbol{\mathcal{K}}_d^m $ and $\boldsymbol{\mathcal{B}}_d^m$ are updated as follows 
\ea{
\label{equ-impedance-adaptation}
\boldsymbol{\mathcal{K}}_d^m = \boldsymbol{\mathcal{K}}_r + \boldsymbol{\mathcal{K}}_s^{m-1}, \quad 
\boldsymbol{\mathcal{B}}_d^m = \boldsymbol{\mathcal{B}}_r + \boldsymbol{\mathcal{B}}_s^{m-1}, 
}  
where $\boldsymbol{\mathcal{K}}_r$ and $\boldsymbol{\mathcal{B}}_r$ are the initial stiffness and damping based on the individual pre-test phase for each human learner and selected reference character. 
As depicted in~\algref{alg-training-procedure}, $L$ initial writing images for the reference character $\bc$ are collected as $\{\boldsymbol{I}_a^l\}_{l=1}^L$. 
The initial writing waypoints are extracted from the collected writing images as $\{\{\bchi_a^{l,n}\}_{n=1}^N\}_{l=1}^L$. 
The mean initial writing waypoints $\overline{\bchi}_a^I$ can be estimated following~\secref{writing-style-evaluation}. 
$\bchi_{\bc}$ and $\overline{\bchi}_a^I$ are aligned with the Dynamic Time Warping~(DTW) , as depicted in \figref{fig:raw_overview}(d). 
$\boldsymbol{\mathcal{K}}_r$ is obtained as 
\ea{
\label{equ-initial-stiffness} 
\boldsymbol{\mathcal{K}}_r = \beta_r |\Delta \overline{\bchi}|, \Delta \overline{\bchi} = \overline{\bchi}_a - \bchi_{\bc}, 
}  
where $\beta_r$ is the coefficient.  
In practice, damping $\boldsymbol{\mathcal{B}}_r$ is obtained from the stiffness $\boldsymbol{\mathcal{B}}_r = 1/2\sqrt{\boldsymbol{\mathcal{K}}_r}$. 

$\boldsymbol{\mathcal{K}}_s^m$ and $\boldsymbol{\mathcal{B}}_s^m$ are the stiffness and damping terms to encourage human engagement. 
$\boldsymbol{\mathcal{K}}_s^0$ and $\boldsymbol{\mathcal{B}}_s^0$ for the first teaching iteration are set to zero. 
Then, $\boldsymbol{\mathcal{K}}_s^m$ and $\boldsymbol{\mathcal{B}}_s^m$ are updated according to the $(m-1)$-th writing performance, as follows 
\ea{ 
\label{equ-impedance-adaptation} 
\setlength\abovedisplayskip{-1pt}
\boldsymbol{\mathcal{K}}_s^m &= \boldsymbol{\mathcal{K}}_s^{m-1} + \beta_{\mathcal{K}} \boldsymbol{\Psi}(\Delta\bchi^{m-1}), 
\boldsymbol{\mathcal{B}}_s^m = 1/2\sqrt{\boldsymbol{\mathcal{K}}_s^m},\\
\boldsymbol{\Psi}(\Delta \bchi^{m-1,j}) &= \frac{\exp{(\alpha\bOmega^{m-1,j} - \Pi_j)} - 1}{\exp{(\alpha\bOmega^{m-1,j} - \Pi_j)} + 1},p=0,1,2, 
} 
where $\boldsymbol{\Psi}(\cdot)$ is an element-wise scalar function~(see Fig.~\ref{fig:raw_overview}(g)). 
$\Delta \bchi^{m-1} = \bchi^{m-1} - \bchi_d^{m-1}$ and $\Delta \bchi^{m-1,j}$ is $j$-th element of $\Delta \bchi^{m-1}$. $\boldsymbol{\Omega}^{m-1,j} = \Delta^2\bchi^{m-1,j} - \Pi_j^2$. 
$\beta_{\mathcal{K}}$ is the coefficient to control the convergence rate, $\alpha$ is a positive scaling scalar. 
$\Pi_j$ is the predefined threshold that determines the region whether to increase the assistance to the human learner. 
In practice, the actual stiffness $\mbK_d^m$ is bounded by the minimal stiffness $\mbK_d^{min}$ and the maximal stiffness $\mbK_d^{max}$, to ensure stable and safe interaction during the robot teaching. 
\begin{table*}[]
\centering
\setlength{\abovecaptionskip}{-0.04cm}   
\begin{tabular}{c}
\hspace{-10pt}\includegraphics[width=1.0\textwidth]{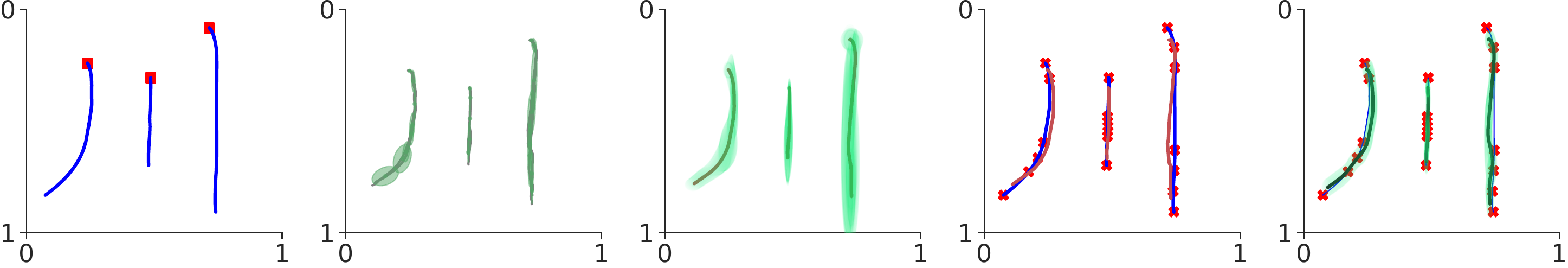} \\  
\hspace{-10pt}\includegraphics[width=1.0\textwidth]{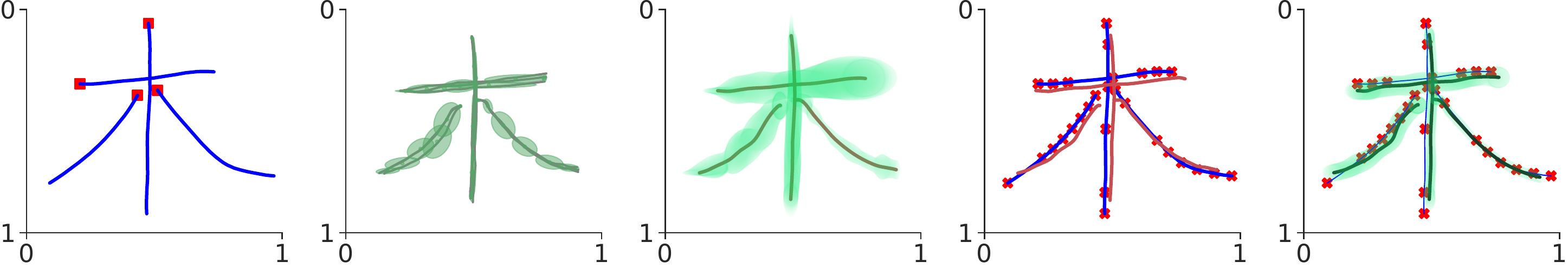} \\ 
(a) \hspace{0.16\textwidth}(b) \hspace{0.17\textwidth}(c) \hspace{0.17\textwidth}(d) \hspace{0.17\textwidth}(e)
\end{tabular} 
\captionof{figure}{Visualization of training waypoints generation for characters $C_3^2$ and $C_4^1$.~(a)~Reference waypoints of each stroke;~(b)~learned GMMs~(green ellipse) from writing waypoints;~(c)~learned style by GMR from writing waypints;~(d)~training via-points;~(e)~generated training waypoints~(black lines) by GMR-GP.} 
\label{tab:vis_training}   
\vspace{-10pt}   
\end{table*}

\vspace{-0.2cm} 
\section{Experiments}\label{exp}  
We conduct human-subject experiments to address the following questions:
\begin{enumerate} 
\item Does \name{} enable human learners to acquire the target skill more efficiently compared to baseline methods? 
\item Can \name{} accurately capture individual human writing styles and provide adaptive guidance to human learners? 
\item Does the learning experience with \name{} encourage active engagement among learners? 
\end{enumerate} 
In our experiments, participants trained with \name{} demonstrated significantly greater improvements in writing quality, particularly in stroke-level similarity to reference characters, compared to baseline methods. The system effectively adapted to individual writing styles, with teaching trajectories progressively aligning with the reference over iterations. Moreover, increased interaction forces reflected greater learner engagement, promoting active participation.
\vspace{-0.3cm}

\begin{table}[!t]    
\normalsize
\centering
\caption{Chinese Characters for Teaching} 
\begin{tabular}{ccccc}
\toprule
$\mbC_{s1}$ & $\mbC_{s2}$ & $\mbC_{s3}$ & $\mbC_{s4}$ & $\mbC_{s5}$ \\
\hline 
\includegraphics[width=0.15\linewidth]{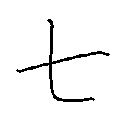}  & 
\includegraphics[width=0.15\linewidth]{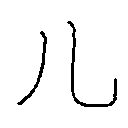} &  
\includegraphics[width=0.15\linewidth]{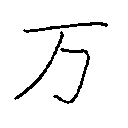}  &  
\includegraphics[width=0.15\linewidth]{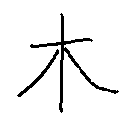}  &  
\includegraphics[width=0.15\linewidth]{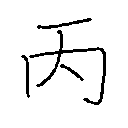} 
\\
\includegraphics[width=0.15\linewidth]{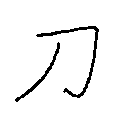}  & 
\includegraphics[width=0.15\linewidth]{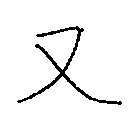}  & 
\includegraphics[width=0.15\linewidth]{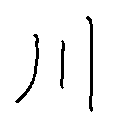}  & 
\includegraphics[width=0.15\linewidth]{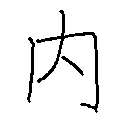}  & 
\includegraphics[width=0.15\linewidth]{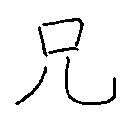} 
\\ 
\includegraphics[width=0.15\linewidth]{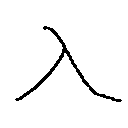}  &  
\includegraphics[width=0.15\linewidth]{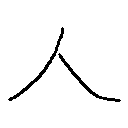}  & 
\includegraphics[width=0.15\linewidth]{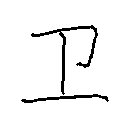}  & 
\includegraphics[width=0.15\linewidth]{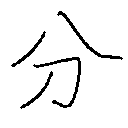}  & 
\includegraphics[width=0.15\linewidth]{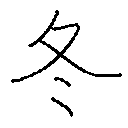}\\
\bottomrule
\end{tabular}
\label{tab:dataset} 
\vspace{-0.5cm} 
\end{table}   
  
\subsection{Experimental Setups} 
\subsubsection{Hardware Setup} 
We use the 7-DoF torque-controlled Franka Emika Research 3~(FR3) robot for all handwriting teaching experiments. As shown in Fig.~\ref{fig:introduction}, the writing workspace measures 350~mm $\times$ 350~mm. 
Interaction forces at the end-effector handle are estimated from joint torques. The reference character~$\bc$ is represented by a 128~$\times$~128 image. During pre-test and evaluation, learners’ handwritten characters are captured by a mounted camera and resized to 128~$\times$~128. 
Reference and writing waypoints are extracted from these images using feature extraction methods. 
During robotic teaching, the learner’s actual writing trajectories are recorded in real time from the end-effector position. 

\subsubsection{Algorithm Setup} 
The robot teacher’s variable impedance control runs at 1000~Hz~($T_s=0.001$~s). 
Stiffness ranges from $\boldsymbol{\mathcal{K}}_d^{min} = [200, 200, 200]$~N/m to $\boldsymbol{\mathcal{K}}_d^{max} = [1200, 1200, 1200]$~N/m. 
Parameters for writing style learning and waypoint generation are $N=200$, $L=3$, $H=5$, and $M=9$. 
Impedance variation parameters are set to $\beta_r = 1000$~N/m$^2$, $\beta_\mathcal{K} = 100$~N/m$^2$, $\Pi_j = 0.05$~m, and $\alpha = 2000$.
\subsubsection{Training Dataset}  
The Chinese character dataset $\mbC$ used in this study is obtained from \cite{chinese-dataset}. We selected 15 distinct characters from the dataset, which includes different numbers of strokes, representing levels of complexity~\cite{luo2024callirewrite}. 
$\mbC_{si}$ represents the subdataset including characters with $i$ strokes. 
$\bc = C_i^j$ represents the $j-$th character with $i$ strokes. 
As depicted in Table~\ref{tab:dataset}, all characters are divided into five groups according to the number of strokes. Each group includes three commonly used Chinese characters, which covers the commonly used strokes in Chinese characters~\cite{lemaignan2016learning,luo2024callirewrite,wang2024rodal}. As depicted in~\figref{tab:vis_training}, we visualized the process for generating the training waypoints of characters $C_3^2$ and $C_4^1$, following the implementation in~\algref{alg-training-procedure}. 

\begin{table}[t]
\centering
\caption{Information of Participants}
\small %
\setlength{\tabcolsep}{13pt} %
\begin{tabular}{ccc}
\toprule
\textbf{Level \#0} & \textbf{Level \#1} & \textbf{Level \#2} \\
\midrule
$P_0^1$~(Indian)   & $P_1^1$~(Singaporean) & $P_2^1$~(Chinese) \\
$P_0^2$~(Indian)   & $P_1^2$~(Singaporean) & $P_2^2$~(Chinese) \\
$P_0^3$~(Turk)  & $P_1^3$~(Malaysian)   & $P_2^3$~(Chinese) \\
$P_0^4$~(Ghanaian) & $P_1^4$~(Malaysian)   & $P_2^4$~(Chinese) \\
$P_0^5$~(Korean)   & $P_1^5$~(Indonesian)  & $P_2^5$~(Chinese) \\
\bottomrule
\end{tabular}
\label{tab:participants}
\vspace{-12pt} %
\end{table}
  
\subsubsection{Participants}  
15 subjects~(5 females and 10 males) were recruited to participate in the robot teaching experiments. All participants were highly educated university students from diverse national backgrounds. 
They were categorized into three levels based on their prior knowledge of Chinese character writing, as summarized in Table~\ref{tab:participants}. 
Each participant is denoted as $P_i^j$, where $i$ represents the level and $j$ represents the subject index within that level. 
\begin{enumerate} 
\item \emph{Level \#0}: No prior knowledge or experience with the characters (e.g., Indian, Turkish, Korean students).  
\item \emph{Level \#1}: Basic familiarity but no regular use (e.g., Singaporean, Malaysian students).  
\item \emph{Level \#2}: Full proficiency with daily usage (e.g., Chinese students).
\end{enumerate} 
\vspace{-0.4cm} 
\subsection{Experimental Protocol} 
The effectiveness of \name{} is further validated by comparing it to two baselines methods: font copy~(FC) and robot guided writing~(RGW). 
For each participant, all 15 Chinese characters in Table~\ref{tab:dataset} are randomly divided into three groups. Each group of characters include five characters with stroke counts ranging from 1 to 5. 
Each participant learns three groups of Chinese characters using different methods for the same writing times in a random order, as follows: 
\begin{enumerate}  
\item \emph{\name{}}: The group of Chinese characters taught using \name{} is referred to as \emph{Group 1}~($G_1$).  
\item \emph{FC}: Participants are only allowed to look at the reference character without any physical correction from the robot. 
The group of characters taught by FC is referred to as \emph{Group 2}~($G_2$). 
\item \emph{RGW}: 
Inspired by robotic physical training~\cite{li2018iterative,yang2022task}, RGW uses maximum stiffness $\boldsymbol{\mathcal{K}}_d^{max}$ to fully guide participants in replicating the reference trajectory without impedance and trajectory variation. It means that the human learner active engagement is not considered. 
The group of characters taught by RGW is referred to as \emph{Group 3}~($G_3$). 
\end{enumerate} 
\begin{figure}[t]
\setlength{\abovecaptionskip}{-0.00cm}  
\centering
\includegraphics[width=1.0\linewidth]{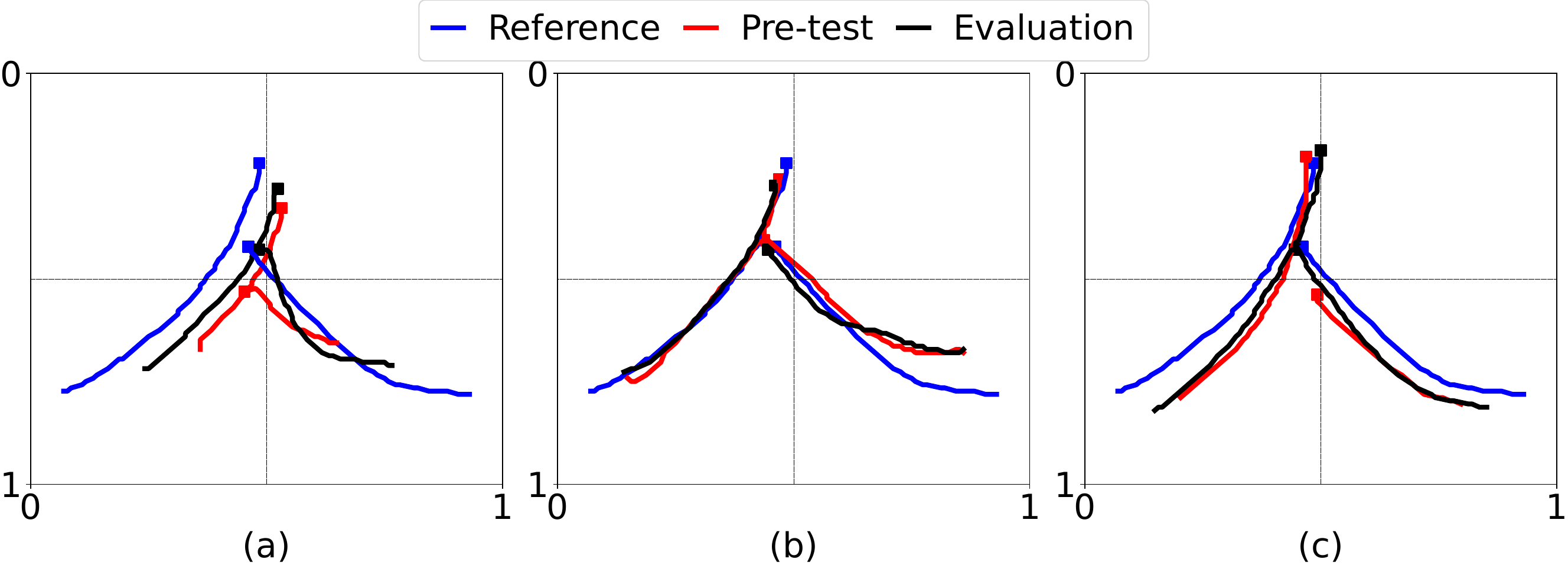} 
\caption{
Writing performance based on $M_1$ for character $C_2^3$ using three different teaching methods. The red lines are the writing waypoints from the pre-test phase and the black lines are the writing waypoints from the evaluation phase.~(a)~FC;~(b)~RGW;~(c)~Ours.} 
\label{fig:m1_ren}  
\vspace{-0.6cm} 
\end{figure}

The experiments for each character is implemented following three phases: 
\begin{enumerate}  
\item \noindent\emph{Pre-test}: 
The pre-test phase is conducted before the teaching phase to assess the initial writing performance of each participant, eliminating the influence of prior knowledge. 
During this phase, participants write the given reference character $L$ times without any physical or visual guidance. 
The individual writing duration $\Delta \bt_{\bc}$ for each given reference character is measured for each participant. 
Then, the initial stiffness and damping for each participant are calculated accordingly. 
\item \noindent\emph{Teaching}: 
During the teaching phase, all participants learn characters in $G_1$ and $G_2$ by grasping the robot handle and observing the actual writing trajectories. 
For characters in $G_3$, all participants first watch a simulated character writing sequence on the screen, after which they attempt to write the character by hand without visual or physical guidance. 
\item \noindent\emph{Evaluation}: 
The evaluation phase is conducted after the teaching phase. Participants write the same reference character $L$ times again, without any reference or guidance.  
\end{enumerate}  
\vspace{-0.3cm}
\subsection{Experimental Results}  
\subsubsection{Writing Performance} 
\begin{figure}[t] 
\setlength{\abovecaptionskip}{-0.00cm}  
\centering
\includegraphics[width=1.0\linewidth]{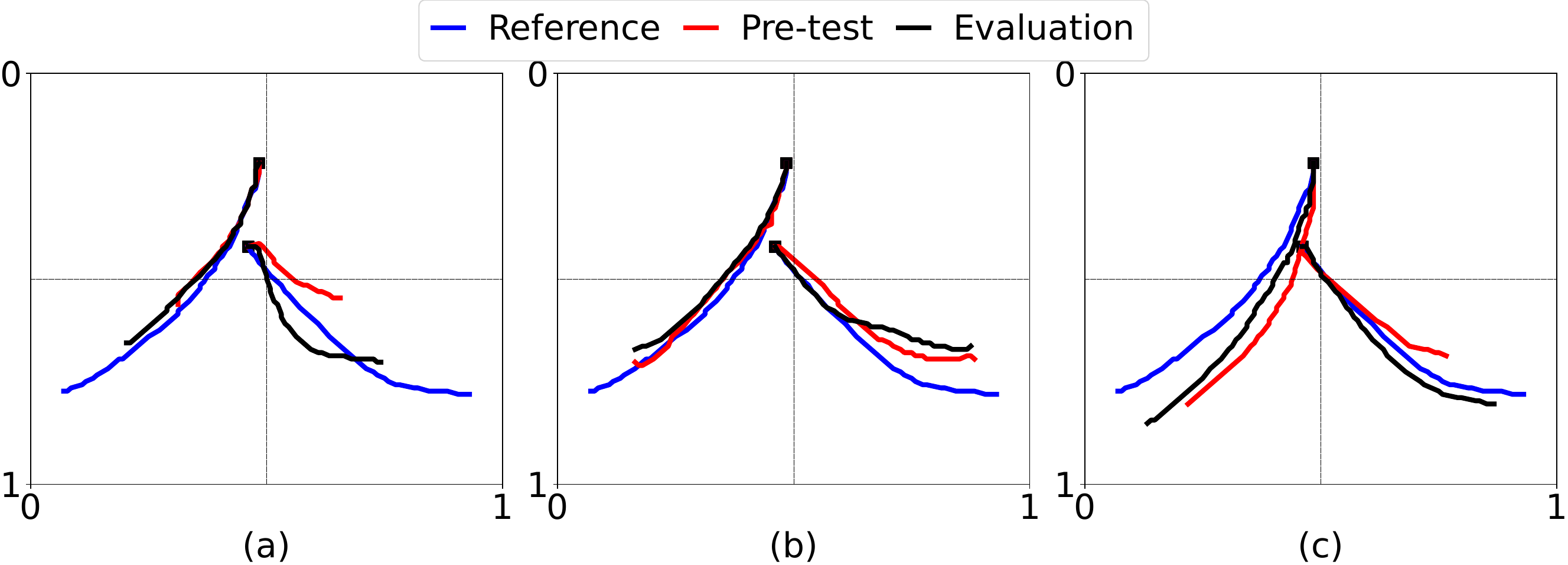}  
\caption{
Writing performance based on $M_2$ of character $C_2^3$ using three different teaching methods. The red lines are the writing waypoints from the pre-test phase and the black lines are the writing waypoints from the evaluation phase.~(a)~FC; (b)~RGW; (c)~Ours.} 
\label{fig:m2_ren}  
\vspace{-0.6cm} 
\end{figure}   
  
\begin{table*}[t]
    \centering
    \setlength{\abovecaptionskip}{-0.02cm}   
    \begin{tabular}{ccc}\hspace{-10pt}
    \includegraphics[width=0.29\linewidth]{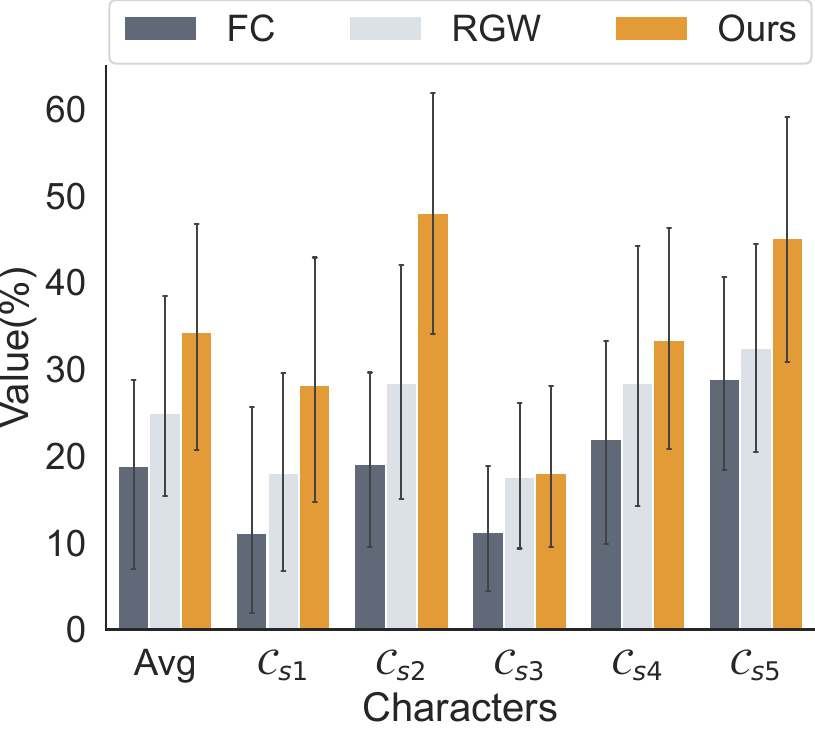}
    \label{fig:sec3-inllustration-framework} & 
    \includegraphics[width=0.29\linewidth]{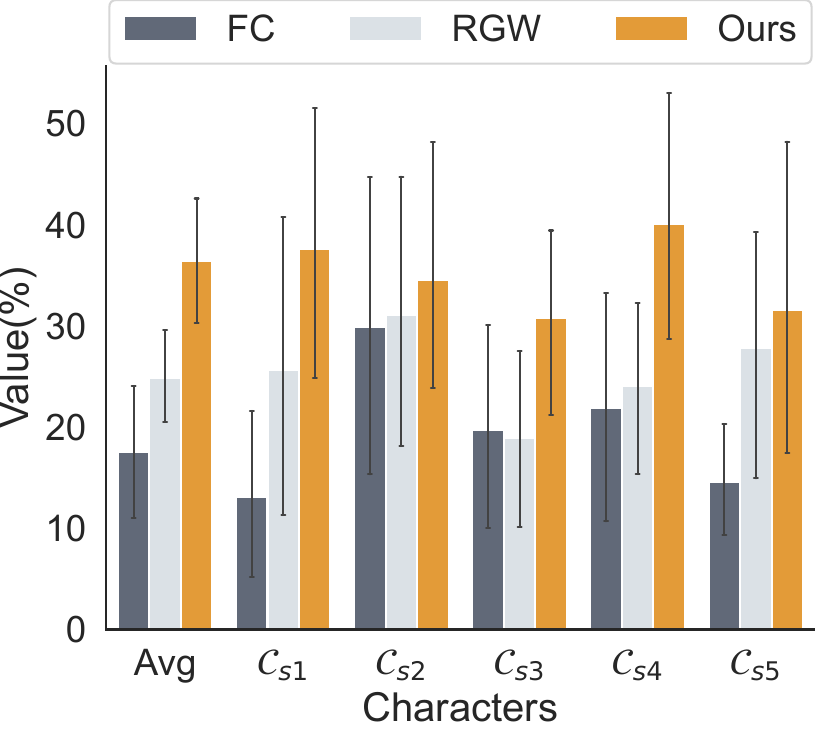}   
    \label{fig:sec2-block-diagram} & 
    \includegraphics[width=0.29\linewidth]{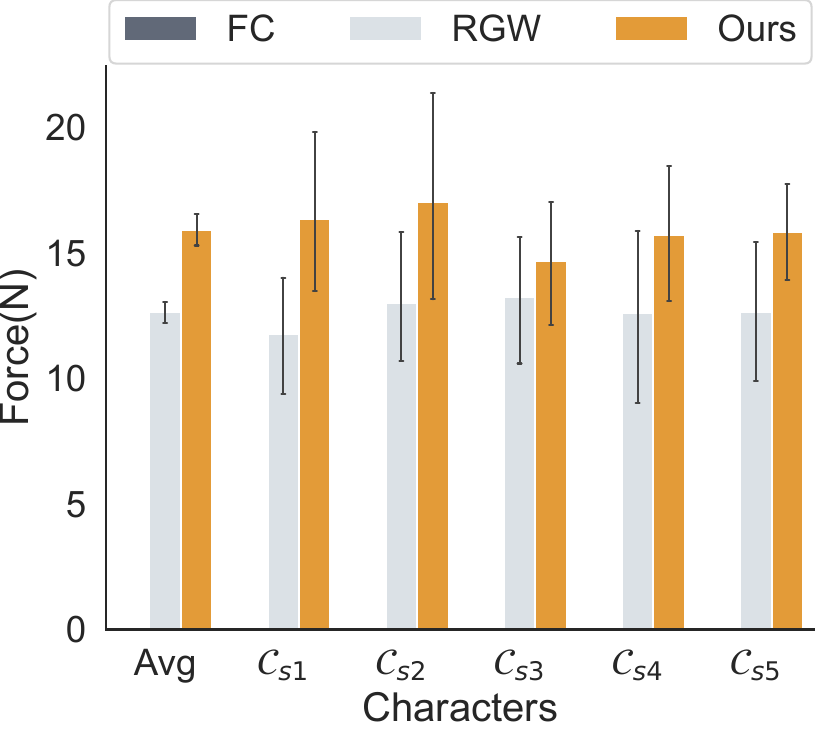}  
    \label{fig:sec2-block-diagram} \\
    (a) & (b) & \hspace{5pt}(c)
    \end{tabular}  
    \captionof{figure}{
    Statistical results of similarity. Improvement and interaction force of three experimental groups across five groups of Chinese characters.  
    (a)~Percentage of similarity improvement based on $M_1$.  
    (b)~Percentage of similarity improvement based on $M_2$.  
    (c)~Average interaction force per waypoint. } 
    \label{tab:efficacy}   
    \vspace{-0.1cm}     
\end{table*}    
  
\begin{table*}[t]
\centering
\setlength{\abovecaptionskip}{-0.02cm}   
\begin{tabular}{ccccc}
\hspace{-5pt}
\includegraphics[width=0.165\textwidth]{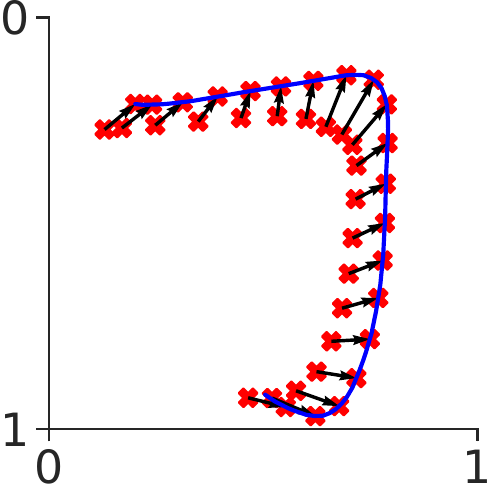}
\label{fig:stiffness}   &  
\includegraphics[width=0.165\textwidth]
{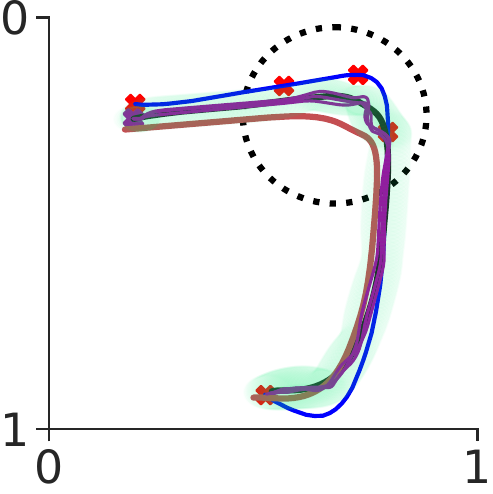}   
\label{fig:sec2-block-diagram} & 
\includegraphics[width=0.165\textwidth]
{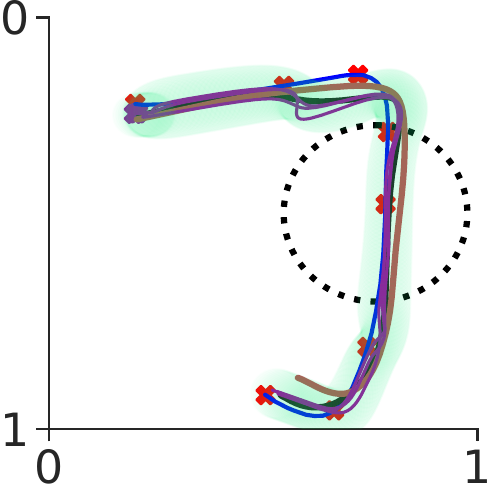} 
\label{fig:sec2-block-diagram} & 
\includegraphics[width=0.165\textwidth]
{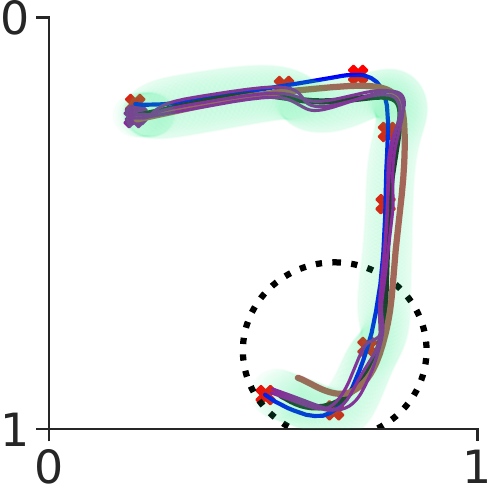} 
\label{fig:sec2-block-diagram} & 
\includegraphics[width=0.165\textwidth]
{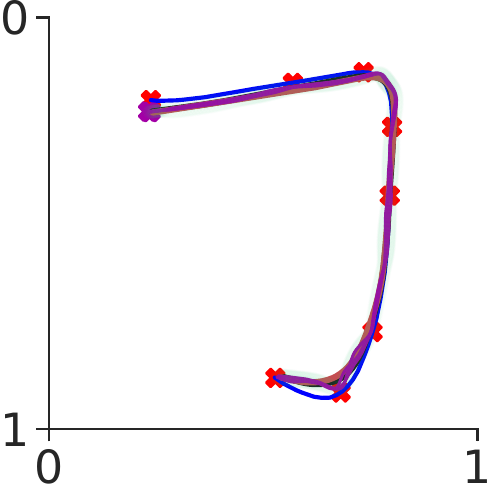} 
\label{fig:sec2-block-diagram} \\ 
(a) & \hspace{3pt}(b) & \hspace{5pt}(c) & \hspace{7pt}(d) & (e)
\end{tabular}  
\captionof{figure}{Visualization of training waypoints changes. 
Blues lines are reference waypoints, scatter red points are training via-points, red lines are learned means of writing waypoints, black lines are the mean of training waypoints, and purple lines are three sampled potential writing waypoints. 
~(a)~Initial reference stiffness generation;
~(b)~$1$-st teaching iteration;
~(c)~$3$-rd teaching iteration;
~(d)~$6$-th teaching iteration;
~(e)~$9$-th teaching iteration.}  
\label{fig:trajectory_fitting_process}  
\vspace{-0.4cm}   
\end{table*} 
  
\begin{table}[t]  
\centering   
\setlength{\tabcolsep}{10pt} %
\captionof{table}{Average Similarity Improvement and Interaction Force.} 
\begin{tabular}{l c c c} 
\toprule
\textbf{Method} & \textbf{Value(\%)}$M_1$ & \textbf{Value(\%)}$M_2$ & \textbf{Force~(N)} \\
\hline 
\vspace{2pt}
FC & 18.79 $\pm$ 12.15 & 17.52 $\pm$ 7.62 & NA\\
RGW & 25.01 $\pm$ 13.044 & 24.84 $\pm$ 5.36 & 12.64 $\pm$ 0.50 \\ 
Ours & \bf{34.30 $\pm$ 14.03} & \bf{36.44 $\pm$ 6.99} & \bf{15.90 $\pm$ 0.77} \\  
\bottomrule
\end{tabular}
\label{tab:avg_table_result}
\vspace{-0.5cm} 
\end{table}   

Two key metrics based on Dynamic Time Warping distance are defined to evaluate the similarity between the reference and written waypoints, using two alignment strategies for comparison. 
First, as shown in \figref{fig:m1_ren}, \emph{Metric 1}~($M_1$) measures global structural similarity by aligning the centers of the written and reference characters, assessing overall shape consistency. 
Second, as shown in \figref{fig:m2_ren}, \emph{Metric 2}~($M_2$) measures stroke-wise similarity by aligning stroke starting points, focusing on individual stroke accuracy. 
Finally, participants’ active engagement is assessed through the interaction force exerted during the writing task.

The statistical comparison of writing results for each subdataset~$\mbC_{si}$ is shown in \figref{tab:efficacy}. 
First, we find that all participants across the three groups have similar initial writing performance. 
The average percentage improvement in similarity for each group and character after training is shown in \figref{tab:efficacy}(a) and (b). 
Table~\ref{tab:avg_table_result} summarizes the average improvement in $M_1$ and $M_2$ across characters. 
Compared to FC and RGW, participants trained with \name{} show significantly greater improvement overall. 
For $M_1$ (global structural similarity), the improvement compared to the baselines is not entirely significant ($p < 0.05$ vs.\ FC, $p \approx 0.09$ vs.\ RGW), suggesting that adapting to individual styles has a marginal effect on global structure. 
That said, this result still underscores the value of physical guidance in helping learners grasp character structure. 
In contrast, for $M_2$ (stroke-wise similarity), \name{} shows significantly better improvement than both FC and RGW ($p < 0.05$ for both), highlighting the importance of adaptive strategies in refining fine-grained stroke details. 

\subsubsection{Adaptation of Training Waypoints} 
To illustrate adaptability during robotic teaching, \figref{fig:trajectory_fitting_process} shows the training waypoints for character $C_2^2$ by participant $P_2^3$. 
In \figref{fig:trajectory_fitting_process}(a), the red scatter points show the mean writing waypoints from the pre-test, which differ substantially from the reference waypoints (blue line). 
This difference is used to generate the initial reference stiffness. 
\figref{fig:trajectory_fitting_process}(b)–(e) show the results of the 1st, 3rd, 6th, and 9th teaching iterations, respectively.

In \figref{fig:trajectory_fitting_process}(b), training via-points~($H=5$) are extracted to enable the human learner to capture the curvature of the reference style in the dashed circle. 
At $3$-rd and $6$-th teaching iterations, as illustrated in \figref{fig:trajectory_fitting_process}(c) and (d), training via-points~($H=8$) are extracted to enable the human learner to focus on different parts of reference style~(as shown in the dashed circles). 
As the robot teaching progresses, the generated training waypoints progressively align with the reference waypoints. 
Notably, by the 6-th teaching iteration (\figref{fig:trajectory_fitting_process}(d)), the human learner demonstrates significantly reduced variability in replicating the reference character, indicating improved writing precision. 
\subsubsection{Active Engagement of Human Learners} 
As depicted in \figref{tab:efficacy}(a) and \figref{tab:efficacy}(b), learners using our method show greater similarity improvement compared to those using RGW.  
To validate the human learner active engagement is crucial for writing skill learning, one indicator of active engagement is the interaction force exerted by the human learner. 
A higher interaction force suggests the greater active participation during the writing skill learning. 
We calculate the average interaction force for each character across all learners. As depicted in \figref{tab:efficacy}(c), the interaction force in $G3$ significantly surpasses that in $G2$ ($p < 0.05$), indicating higher human engagement during robot teaching, which aids in improving writing skills. In this paper, we select the same parameters for impedance variation function, which can also be optimized for each learner to achieve higher learning efficiency. 
\begin{table}[t] 
\setlength{\abovecaptionskip}{-0.02cm}   
\centering 
\begin{tabular}{ccc}\hspace{-10pt}
\includegraphics[width=0.47\linewidth]{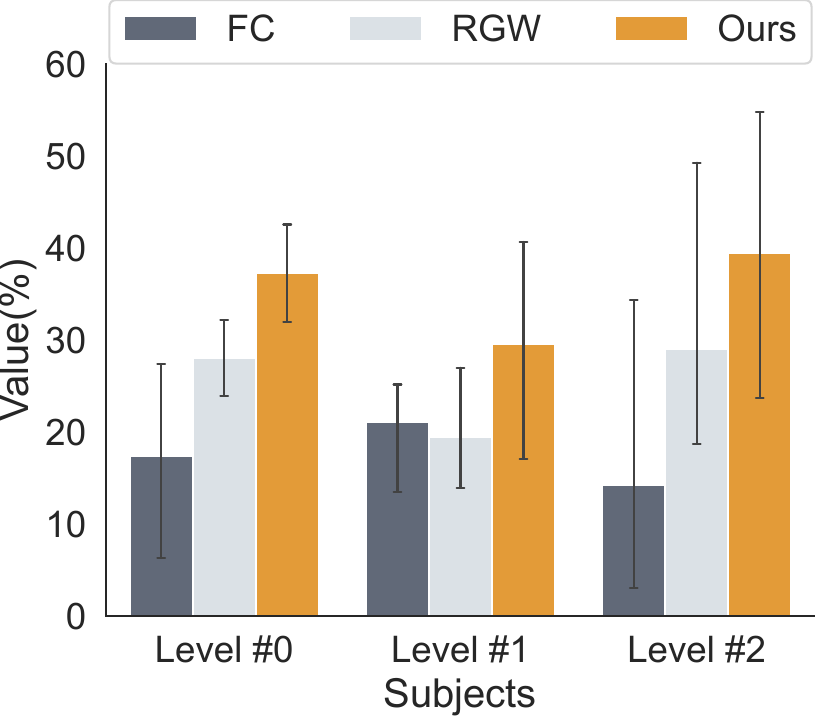}
\label{fig:different-levels-metric-1} &  
\includegraphics[width=0.47\linewidth]{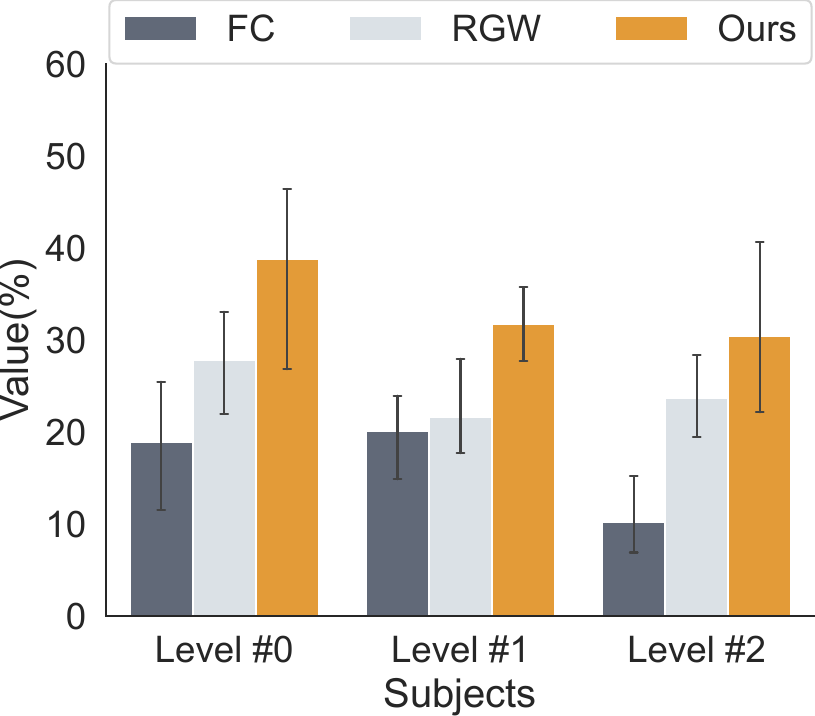}   
\label{fig:different-levels-metric-2}\\ 
(a) & (b) 
\end{tabular}  
\captionof{figure}{
Similarity improvement comparison across all characters for participants with varying levels of prior writing experience.~(a)~Percentage of similarity improvement based on $M_1$. 
(b)~Percentage of similarity improvement based on $M_2$. 
} 
\label{fig:comparison-different-levels}  
\vspace{-0.7cm}     
\end{table}    

\vspace{-0.2cm}   
\subsection{Discussion} 
We execute post-experiment survey to assess perceived improvement of human learners' in writing skills. 
Despite $G3$ showing quantitative performance advantages, learners do not report increased confidence in better writing skills compared with $G2$, consistent with previous research~\cite{Tian2023TowardsMA}. 
This contradiction leads to two hypotheses: \textit{1}) Lack of visual signal integration: human learners may need clear visual signals in addition to physical ones for better understanding. \textit{2}) Failure to emulate human instructors: due to differences in morphology and the challenge of replicating real teaching scenarios, full emulation is difficult. These hypotheses guide future robot teaching system development, emphasizing the importance of multi-modal signals (e.g., language and visual) for greater effectiveness and generality. 
In this paper, the effectiveness is mainly validated by highly educated human users. 
As shown in Fig.~\ref{fig:comparison-different-levels}, human users with different prior writing experiences have tested the performance. 
The participants with level \#2 tend to focus on capturing the global structural similarity, showing improvement in overall similarity when evaluated with metric $M_1$ compared to $M_2$. 
In contrast, participants with level \#1 prior experience primarily focus on stroke-wise improvement, exhibiting lower variance in the similarity based on $M_2$ compared to $M_1$.  
These individuals are generally familiar with the structural layout of the characters but lack sufficient practice in writing them fluently.  
However, no clear trend or consistent conclusion can be drawn for participants with level \#0 prior experience, as their performance does not exhibit a distinct pattern. 
In the future, the human users at different ages will be invited to test the effectiveness. Other characters dataset will be employed to test the effectiveness.

\vspace{-0.1cm}
\section{Conclusion}\label{clu}  
In this work, we introduce \name{}, a robot teaching system in which robots assume the role of instructors, guiding humans in character writing through adaptive physical interactions. Results from human-subject experiments demonstrate the effectiveness of \name{}. Moreover, we provide evidence of how \name{} customizes its teaching approach to meet the unique needs of individual learners, leading to enhanced overall engagement and effectiveness. The potential of robot teaching presents exciting opportunities for scaling up education and providing learning opportunities to many, even in the absence of human teachers. 

For broader applications of physical robot teaching, our findings suggest that effective teaching of physical skills requires personalized guidance, dynamic adjustment of assistance levels, and the integration of physical guidance with multimodal feedback such as visual or verbal cues.

\end{CJK*}

\small 
\bibliographystyle{IEEEtran} 
\bibliography{
reference/robot_teach_writing.bib,
reference/robot_teaching.bib,
reference/trajectory_adaptation.bib
} 
\end{document}